\documentclass{article}

\usepackage[preprint]{neurips_2024}
\usepackage[usenames,dvipsnames]{xcolor}         

\usepackage[colorlinks=True,citecolor=ForestGreen]{hyperref}       

\usepackage[utf8]{inputenc} 
\usepackage[T1]{fontenc}    
\usepackage{hyperref}       
\usepackage{url}            
\usepackage{booktabs}       
\usepackage{amsfonts}       
\usepackage{nicefrac}       
\usepackage{microtype}      
\usepackage{xcolor}         

\usepackage{subcaption}
\usepackage{algorithm}
\usepackage{algorithmic}

\usepackage{amsmath,amsfonts,bm,amssymb}
\usepackage{amsthm}
\usepackage{mathtools}
\usepackage{xcolor}
\usepackage{nicefrac}       

\usepackage[noabbrev,capitalize,nameinlink]{cleveref}

\allowdisplaybreaks

{}

\makeatletter
\newtheorem*{rep@theorem}{\rep@title}
\newcommand{\newreptheorem}[2]{%
\newenvironment{rep#1}[1]{%
 \def\rep@title{#2 \ref{##1}}%
 \begin{rep@theorem}}%
 {\end{rep@theorem}}}
\makeatother

\newreptheorem{theorem}{Theorem}
\newreptheorem{lemma}{Lemma}
\newreptheorem{prop}{Proposition}
\newreptheorem{corollary}{Corollary}
\newreptheorem{definition}{Definition}

\newenvironment{itemize*}%
{\begin{itemize}[leftmargin=*,topsep=0pt]%
		\setlength{\itemsep}{0pt}%
		\setlength{\parskip}{0pt}}%
	{\end{itemize}}
\newenvironment{enumerate*}%
{\begin{enumerate}[leftmargin=*,topsep=0pt]%
		\setlength{\itemsep}{0pt}%
		\setlength{\parskip}{0pt}}%
	{\end{enumerate}}

\def\eqref#1{equation~\ref{#1}}

\def\1{\bm{1}}

\DeclareMathAlphabet{\mathsfit}{\encodingdefault}{\sfdefault}{m}{sl}
\SetMathAlphabet{\mathsfit}{bold}{\encodingdefault}{\sfdefault}{bx}{n}

\def\gF{{\mathcal{F}}}

\usepackage{tabularx}
\usepackage{booktabs}
\usepackage{multirow}
\usepackage{enumitem}
\usepackage{bbm}
\usepackage{enumitem}
\usepackage{manfnt}
\usepackage{comment}

\usepackage[textsize=tiny]{todonotes}

\title{Landscape-Aware Growing:\\The Power of a Little LAG}

\author{%
  \{Stefani Karp\thanks{Joint first authorship. Either author can be cited first.}, ~~~Nikunj Saunshi$^*$\}\\
  Google Research\\
  \texttt{stefanik@google.com, nsaunshi@google.com}\\
  \And
   Sobhan Miryoosefi \\
   Google Research\\
   \texttt{miryoosefi@google.com} \\
   \And
   Sashank J. Reddi\\
   Google Research\\
   \texttt{sashank@google.com} \\
   \And
   Sanjiv Kumar\\
   Google Research\\
   \texttt{sanjivk@google.com} \\
}

\begin{document}

\maketitle

\begin{abstract}
Recently, there has been increasing interest in efficient pretraining paradigms for training Transformer-based models. Several recent approaches use smaller models to initialize larger models in order to save computation (e.g., stacking and fusion). In this work, we study the fundamental question of how to select the best growing strategy from a given pool of growing strategies. Prior works have extensively focused on loss- and/or function-preserving behavior at initialization or simply performance at the end of training. Instead, we identify that behavior at initialization can be misleading as a predictor of final performance and present an alternative perspective based on early training dynamics, which we call ``landscape-aware growing (LAG)''. We perform extensive analysis of correlation of the final performance with performance in the initial steps of training and find early and more accurate predictions of the optimal growing strategy (i.e., with only a small ``lag'' after initialization). This perspective also motivates an adaptive strategy for gradual stacking.
\end{abstract}

\section{Introduction}
\label{sec:intro}

\looseness-1Large language models with hundreds of billions of parameters have changed the landscape of NLP and AI, owing to incredible emergent properties arising at scale.
Training large models, however, requires a lot of time and computational resources, thus necessitating the development of efficient training paradigms.
One way to save resources is to design more efficient pretraining algorithms which require fewer resources and less wallclock time to train large models.
Besides the classical approach of developing better optimization algorithms, another paradigm that has recently been gaining popularity is knowledge transfer via {\em growing of models}.
Rather than training a large model from scratch, the idea is to use a much smaller existing pretrained model to initialize the parameters of the larger model \citep{chen2015net2net,wang2022learning}.
This has been shown to accelerate training of large models compared to training from random initialization.
Recently, such ideas were used to train a 100B-parameter-scale model by growing from a 16B-parameter model \citep{li2023flm}.

\looseness-1A related paradigm is that of stagewise training, where the goal is to train a target model by gradually growing its size. Methods like progressive and gradual stacking \citep{gong2019efficient,reddi2023efficient} have shown great success in efficient stagewise training of BERT models by stacking layers from the previous stage to initialize the next stage. 
All of these approaches for growing and stacking have a crucial design choice of the {\em growth operator} that is used to  initialize the larger model from a smaller model. In the context of growing in depth, the fundamental design question becomes: {\em how should we use an $L$-layer model to grow into an $(L+k)$-layer model for further training?}

\looseness-1Numerous strategies have been proposed in prior work for growing in depth as well as width (see Section~\ref{sec:related} for a literature survey).
These strategies are largely heuristic or based on the principle of {\em loss or function preservation} -- a growing strategy should ensure that the loss value/functional behavior of the grown model is the same or very similar to that of the smaller model used to initialize it.
At first sight this seems like a desirable property, because we already know that the small model is a decent pretrained model.
Thus, growth operators that maintain the loss value or functional properties will guarantee a good initialization for the larger model, which can be further improved upon by continued training.
While this seems intuitive, to the best of our knowledge there is no systematic study of the efficacy of this approach.
Thus, our paper asks and studies the following questions:
\looseness-1\begin{center}
    {\em What are good guiding principles to select the best growing strategy?\\
    Is loss/function preservation a good heuristic?}
\end{center}
\looseness-1In this work, we study the above question in the context of growing in depth\footnote{Note that one could also grow models in width, but for this paper we restrict our attention to depth-wise growing.}.
In particular, through extensive empirical analysis, we argue that loss preservation is not necessarily a good strategy. Instead we identify that the training dynamics and {\em landscape properties} afforded by an initialization play a much bigger role in the success of growing.
In this context, we highlight the following contributions:
\begin{itemize}
    \item For a pool of depth-growing strategies inspired by prior work, we conduct an extensive empirical analysis and find that the loss value at initialization does not correlate well with the final performance of the model (i.e., Pearson correlation of -0.51 and Spearman correlation of -0.42).
    \item Instead, we propose an alternative view based on the landscape induced by an initialization through the following key observation: while initial loss can be misleading, the loss after a relatively small number of steps (roughly 5000) correlates very strongly with the final performance (i.e., Pearson correlation of 0.98 and Spearman correlation of 0.99).
    \item We take this a step further and find that good predictions for the best strategy can be made even earlier, after a few hundred steps, and that this corresponds with a measurable phase transition.
    \item 
    Based on the above empirical observation, we propose a selection strategy called Landscape-Aware Growing (LAG).
    We validate our hypothesis by testing LAG on 1-stage growing for both BERT and UL2 pretraining. LAG is shown to recover a strategy that is very close to the optimal strategy in hindsight and is also better than many previously considered static strategies.
    Furthermore, we apply LAG to the setting of gradual stacking by applying LAG to each stacking stage. This improves BERT pretraining loss compared to vanilla gradual stacking, thus further validating the efficacy of LAG.
\end{itemize}

\section{Problem setup: growing and stacking}
\label{sec:problem_setup}

\looseness-1We describe the problem of growing in general and how it can be applied to stacking. This section also sets up the notation for the rest of the paper.

\subsection{Growing}
\label{sec:setup_growing}

\begin{figure}
    \centering
    \begin{subfigure}{0.27\textwidth}
    \centering
    \includegraphics[width=\textwidth]{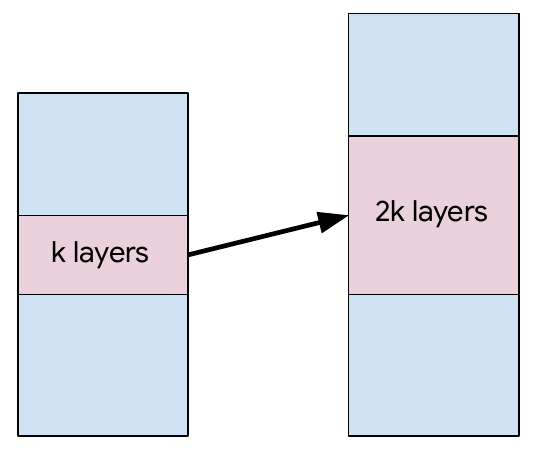}
    \end{subfigure}\hspace{2em}
    \begin{subfigure}{0.27\textwidth}
    \centering
    \includegraphics[width=\textwidth]{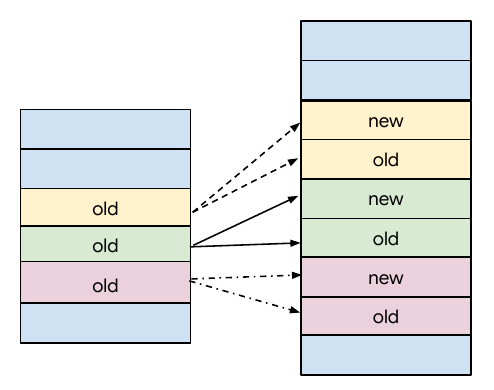} 
    \end{subfigure}\hspace{2em}
    \begin{subfigure}{0.27\textwidth}
    \centering
    \includegraphics[width=\textwidth]{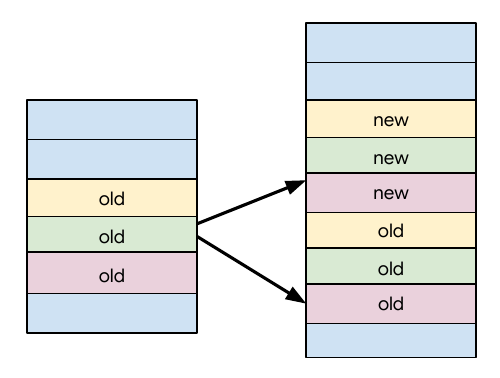}
    \end{subfigure}
    \caption{Illustration of growing a network. Left: Generic growth of $k$ layers into $2k$ layers. Middle: Example with $L=6,k=3,i=2,b=1$, parameter duplication (interleaving). Right: Example with $L=6,k=3,i=2,b=3$, parameter duplication (single-block copying).}
    \label{fig:growing_illustration}
\end{figure}

\looseness-1The basic idea of growing is to use one pretrained model (typically a smaller one) to construct an initialization for another model.
This growing step can be formalized as a growth operator that maps parameters from a smaller architecture class $\gF_1$ to a larger one $\gF_2$.
We let $G: \gF_1 \rightarrow \gF_2$ denote the growth operator.
Given a pretrained checkpoint for a smaller model $M_1$, the grown checkpoint $M_2 = G(M_1)$ expresses every parameter in the new checkpoint as a function of parameters of the old checkpoint.
For Transformer-based architectures, the small and large model classes $\gF_1$ and $\gF_2$ can potentially vary in either the depth dimension (number of layers) or in width (model dimension, number of attention heads).
In this work, for simplicity of analysis, we focus on growing in the depth dimension, i.e., $\gF_1$ and $\gF_2$ only vary in the depth of the model.
However, we note that the ideas being discussed in the paper are general and also apply to other growth dimensions.

\looseness-1We consider growing an $L$-layer network into an $(L+k)$-layer network in the following manner: choose one set of $k$ consecutive layers and grow it into $2k$ consecutive layers (in various ways).
Motivated by prior work on depth growing~\citep{gong2019efficient,reddi2023efficient,wang2022learning}, we consider the following design space, parameterized by \emph{index}, \emph{block size}, and \emph{initialization scheme}:

\begin{itemize}
    \item \textit{Index ($i$)}: When choosing one set of $k$ consecutive layers in the original $L$-layer network, where should this set of $k$ layers begin? Throughout, we will let $i$ denote the start index for this set of $k$ layers (1-indexed), so that the feasible set of values for $i$ is $\{1,2,\dots,L-k+1\}$.
    \item \textit{Block size ($b$)}: How should we divide $k$ into smaller blocks when growing? For simplicity, we divide $k$ into equal-sized blocks and consider all valid block sizes $b$ such that $k$ is divisible by $b$. At one extreme, when $b=1$, new and old layers are interleaved. At the other extreme, when $b=k$, the $k$ old layers end up consecutive and the $k$ new layers end up consecutive in the $(L+k)$-layer network.
    \item \textit{Initialization scheme}: How should the new layers be initialized? We consider both options of random initialization and parameter duplication of the old layers.
\end{itemize}

\looseness-1To make this more concrete with an example, let us consider a 6-layer network that we wish to grow to 9 layers. In this case, $L=6$ and $k=3$. The feasible set of start indices is $\{1,2,3,4\}$, and the feasible set of block sizes is $\{1,3\}$. To help illustrate how the growth operators work, some examples in the design space are as follows, with the new layers indicated in bold.

\begin{enumerate}
    \item $i=2, b=1$, duplication: $[1,2,3,4,5,6]\rightarrow[1,2,\textbf{2},3,\textbf{3},4,\textbf{4},5,6]$
    \item $i=2, b=3$, duplication: $[1,2,3,4,5,6]\rightarrow[1,2,3,4,\textbf{2},\textbf{3},\textbf{4},5,6]$
    \item $i=2, b=1$, random: $[1,2,3,4,5,6]\rightarrow[1,2,\textbf{\text{random}},3,\textbf{\text{random}},4,\textbf{\text{random}},5,6]$
    \item $i=2, b=3$, random: $[1,2,3,4,5,6]\rightarrow[1,2,3,4,\textbf{\text{random}},\textbf{\text{random}},\textbf{\text{random}},5,6]$
\end{enumerate}

\looseness-1Figure~\ref{fig:growing_illustration} (middle, right) illustrates rows 1, 2 above. 
After the new layers are added, all $L+k$ layers of the resulting model are jointly trained.

\subsection{Stacking as iterated growing}
\label{sec:setup_stacking}

\looseness-1Above, we discuss a single-step growing operation to transition from $L$ to $L+k$ layers.
Throughout, as in \cite{gong2019efficient} and \cite{reddi2023efficient}, we use ``stacking'' to refer to the iterated application of this ``grow, then train'' strategy, where training starts with a shallow model, and at the end of each stage the model depth is grown by a certain amount until the desired depth is achieved.
Thus, any growth operator can be converted into a corresponding stage-wise pretraining approach.
In this work, we consider the gradual post-stacking framework from \cite{reddi2023efficient}, corresponding to the repeated application of growing with start index $L-b+1$ and block size $b$. We use this perspective to extend our growing results in Section~\ref{sec:bert_exps} to stacking in Section~\ref{sec:stacking}.

\section{Understanding growing in depth}
\label{sec:bert_exps}

\subsection{Pitfalls of loss-preservation-based growing}
\label{sec:bert_exps_pitfalls}

\looseness-1As discussed in Section~\ref{sec:intro}, a common idea for growing in prior work is based on loss or function preservation as a guiding principle -- a growth operator is constructed such that it maintains the same loss value or functional behavior as the original smaller model. The intuition is that this can provide a good initialization for the model in terms of the loss, and that hopefully translates to the final performance.
Our work challenges the idea of loss preservation for growing in depth. To put this to the test, we consider a list of depth-growing strategies inspired by prior work (discussed in Section~\ref{sec:setup_growing}) and measure the correlation between the initial and final loss values across these strategies.

\looseness-1\textbf{Growing BERT.}~~~We begin by pretraining BERT-\textsc{Base} for 500,000 steps. See Appendix~\ref{app:training_details} for training details. After 500,000 steps, we grow the model from 12 layers to 16 layers. We consider the abstract design space formalized in Section~\ref{sec:setup_growing}, instantiated in this particular setting. Specifically, we consider indices 0, 2, 4, 6, and 8 (skipping odd indices simply due to compute limitations -- odd indices are equally valid choices), block sizes 1, 2, and 4, and initialization schemes \emph{random} and \emph{parameter duplication}. This results in a search space of 30 different growth operators. 

\looseness-1\textbf{Correlation of initial loss with final loss.}~~~For each of the 30 growth operators $G$, we apply $G$ to the 12-layer model $M_1$ to initialize a 16-layer model $M_2$. We then continue training $M_2$ for 100,000 steps, ensuring that each 16-layer model sees data in the same order, to avoid artifacts of data order as best as possible. We load the optimizer state from the previous stage and maintain a constant learning rate of 0.0001. We measure the validation loss upon initialization of $M_2$ (i.e., at step 500,000), as well as after 100,000 steps of training $M_2$ (i.e., at step 600,000). In Figure~\ref{fig:correlation_12to16} (left), for each growth operator, we plot the validation loss at step 600,000 vs. the validation loss at step 500,000. Visually, we can see that the loss immediately after applying the growth operator is not well-correlated with the loss after continued training for 100,000 steps. Numerically, the Pearson correlation between the validation losses at step 600,000 and the validation losses at step 500,000 is -0.515, and the corresponding Spearman correlation is -0.418.

\looseness-1At step 500,000, the 12-layer model has a validation loss of 1.823. Therefore, although our search space does not include pure loss-preserving growth operators, some of the growth operators do come quite close (e.g., loss approximately 1.825 vs. 1.823), and others only raise the validation loss a bit in comparison with other growth operators in the search space. From this, we can see that approximate loss preservation does not appear to correlate with final performance.

\looseness-1\textbf{Implications.}~~~Given that loss preservation is not a good heuristic to predict the final model performance, is there another approach that is more predictive?
In this rest of the section, we provide a series of empirical analyses suggesting that the early loss landscape can provide a much stronger indicator of the final loss.

\subsection{Strong correlation with final performance emerges early}
\label{sec:early_correlation}

\looseness-1While the initial loss is not very predictive of final performance, we make a surprising discovery that the loss after a few steps of training can be very highly predictive.

\looseness-1\textit{Hypothesis 1}: Loss after some steps of training strongly correlates with the final loss.

\begin{figure}[!tbp]
    \centering
    \begin{subfigure}{0.49\textwidth}
    \centering
    \includegraphics[width=0.95\textwidth]{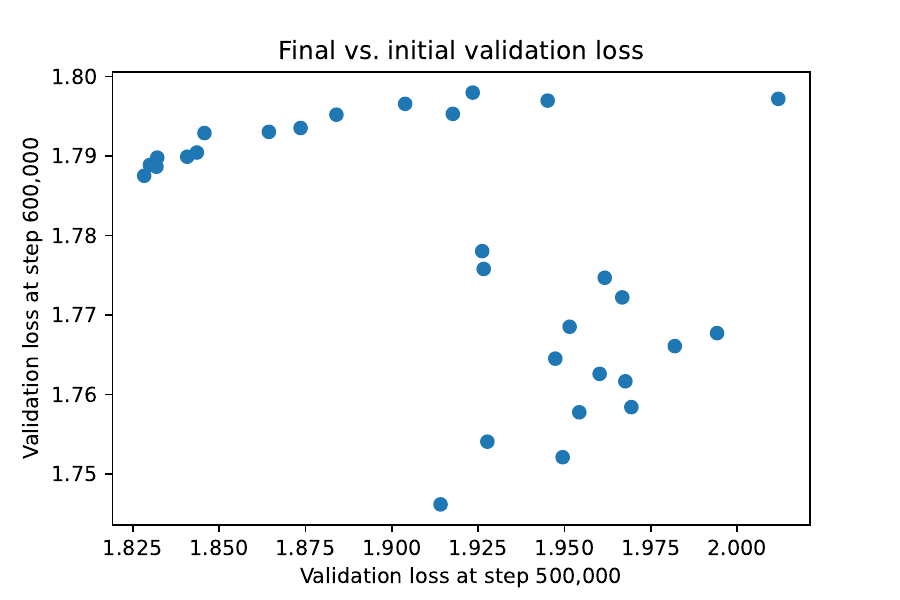}
    \label{fig:600_vs_500_12to16}
    \end{subfigure}\hfill
    \begin{subfigure}{0.49\textwidth}
    \centering
    \includegraphics[width=0.95\textwidth]{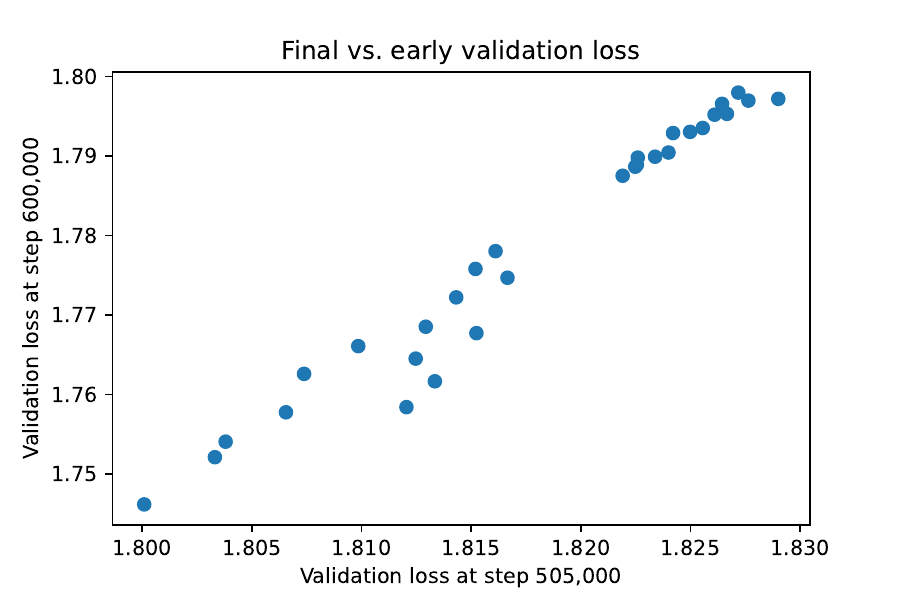}
    \label{fig:600_vs_505_12to16}
    \end{subfigure}\hfill
    \caption{\looseness-1Growing a 12-layer BERT model at step 500,000 into a 16-layer BERT model and then training the larger model for 100K steps. Correlation (in validation loss) between (left) the loss at 600K steps and the loss immediately upon growing (i.e., without any training) and (right) the loss at 600K steps and the loss at 505K steps (i.e., after 5,000 steps of training the larger model).}
    \label{fig:correlation_12to16}
\end{figure}

\looseness-1For each growth operator $G$, we measure the validation loss at step 505,000 (i.e., after 5,000 steps of training $M_2$). In Figure~\ref{fig:correlation_12to16} (right), for each growth operator, we plot the validation loss at step 600,000 vs. the validation loss at step 505,000. Visually, we can see that the loss after 5,000 steps \emph{is highly correlated} with the loss at step 600,000. Numerically, the Pearson correlation between the validation losses at step 600,000 and the validation losses at step 505,000 is 0.982, and the corresponding Spearman correlation is 0.986. In other words, we see that the the final order of the various growing strategies has largely already emerged within the first 5,000 steps. Thus, in this BERT setting, we have strong support for Hypothesis 1.

\looseness-1\textbf{The early loss landscape perspective.}~~~In general, it can be difficult to predict final performance based on initial \textit{or} early performance. However, we posit that, when \textit{growing} an already-trained model, the early loss landscape has particularly nice properties amenable to early prediction. One possible mental model is as follows. Upon growing, the network enters an unstable state induced by the addition of its new layers. However, since the network is already largely trained, it is able to adapt quickly to use its new layers, resulting in a fast drop in the loss. During this initial fast adaptation phase, the overall ordering of various growing strategies is still unstable. However, once this fast adaptation to the new layers is complete, the loss curves and, crucially, their relative orderings enter a more stable phase. This can be summarized as:

\looseness-1\emph{Phase 1: At initialization, look for a nearby point in the loss landscape that is much better adapted to using the new layers (resulting in a rapid drop in the loss).}

\looseness-1\emph{Phase 2: Continue training from this adapted initialization (now in a slower, more predictable manner).}

\looseness-1This perspective suggests that the initial loss is not the only factor in determining the final performance of a particular growth operator. Rather, how the growth operator influences the network's loss landscape near its new initialization is also crucial in determining its final performance. We call this perspective \emph{landscape-aware growing}.

\subsection{Prediction within several hundred steps}
\label{sec:bert_exps_hundredsteps}

\begin{figure}[!tbp]
    \centering
    \begin{subfigure}{0.49\textwidth}
    \includegraphics[width=\textwidth]{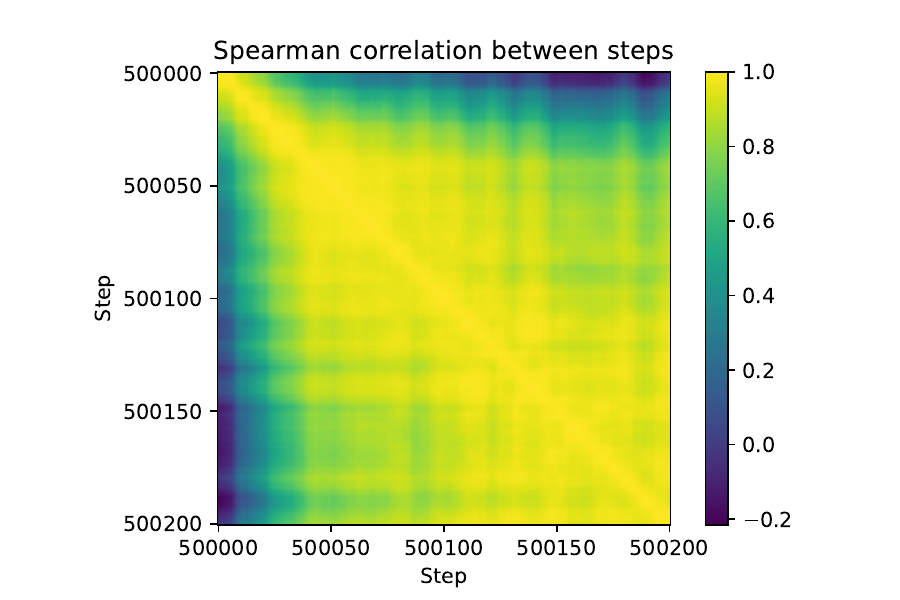}
    \end{subfigure}
    \begin{subfigure}{0.49\textwidth}
    \includegraphics[width=\textwidth]{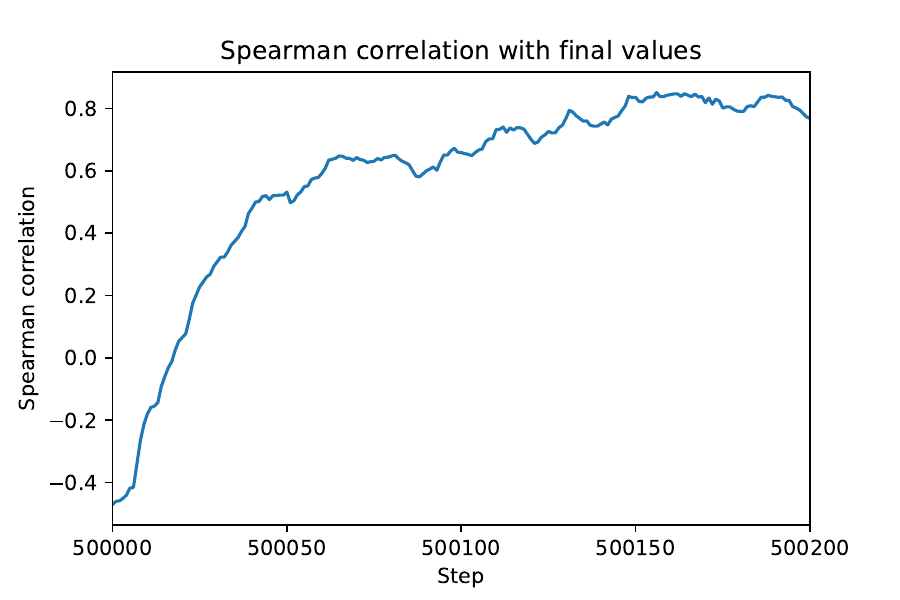}
    \end{subfigure}
    \begin{subfigure}{0.49\textwidth}
    \includegraphics[width=\textwidth]{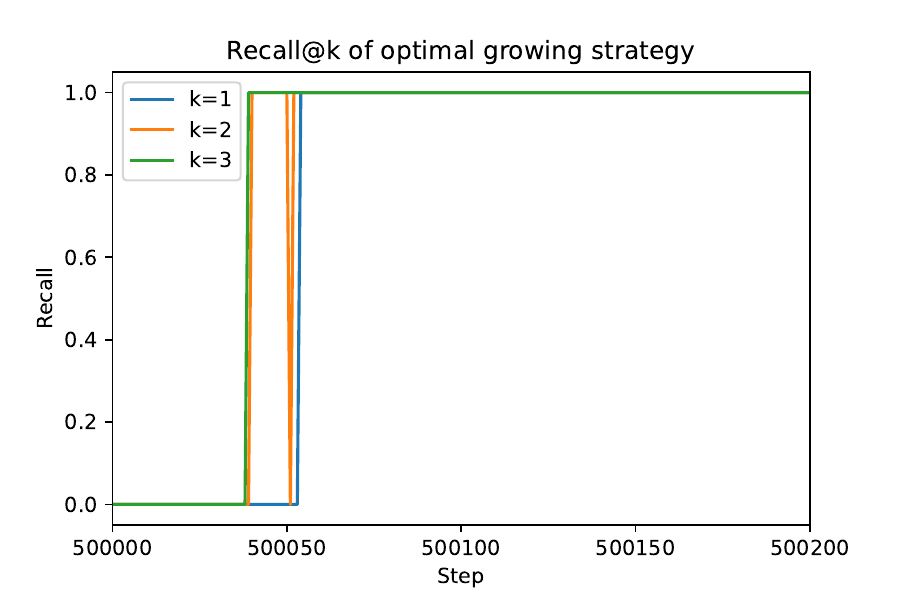}
    \end{subfigure}
    \begin{subfigure}{0.49\textwidth}
    \includegraphics[width=\textwidth]{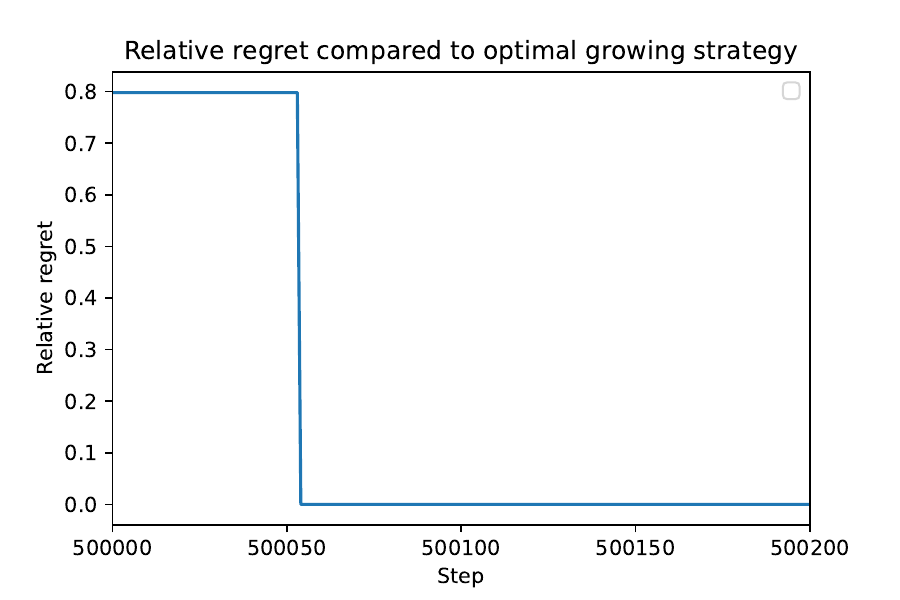}
    \end{subfigure}
    \caption{Growing BERT from 12 to 16 layers: zooming in on steps 500,000 through 500,200. Spearman correlation heatmap (top left), Spearman correlation with final values (top right), Recall@$k$ (bottom left), Relative regret (bottom right). See below for details on how these plots were constructed. For all plots, at each step, the validation loss is first averaged over a window of 11 steps (centered at the step in question) to help smooth out noise.}
    \label{fig:bert_12_to_16_deep_dive}
\end{figure}

\looseness-1Based on our insights in Section~\ref{sec:early_correlation}, we ask: \emph{How early does good prediction become possible?} We zoom in on the first 200 steps of training after growing (i.e., steps 500,000 to 500,200) and observe several interesting properties.

\looseness-1\textbf{Self-correlation heatmap.}~~~For every pair of steps $i,j \in \{500000,\dots,500200\} \times \{500000,\dots,500200\}$, we compute the Spearman correlation between the validation losses at step $i$ and the validation losses at step $j$. Figure~\ref{fig:bert_12_to_16_deep_dive} (top left) displays this as a correlation heatmap. In this heatmap, we can see a clear phase transition between the earliest few steps of training (whose validation losses do not correlate well with the losses at step 500,200) and the remainder of the first 200 steps, which have strong correlation amongst themselves. This phase transition between the earliest few steps of training and the remainder of the first 200 steps seems to occur roughly around step 500,050.

\looseness-1\textbf{Correlation with final performance.}~~~For every step $i \in \{500000,\dots,500200\}$, we compute the Spearman correlation between the validation losses at step $i$ and the validation losses at step 600,000. We observe that the Spearman correlation rises rapidly within the first 200 steps, from less than -0.4 to approximately 0.8. This indicates that strong correlation with final performance emerges very early on in training, providing some support for the conjecture that the phase transition observed above is, in fact, a transition into a more stable phase with high predictive power with respect to final performance.

\looseness-1\textbf{Measuring Recall@k.}~~~Practically speaking, however, we do not need to predict the \textit{full ranking} of strategies. In fact, predicting the exact ordering among the worst-performing strategies has very limited utility. Rather, a practitioner could exploit the above insights by pruning their search space down to just a few top configurations. Then, one practically-relevant question is: Was the best growing strategy within this pruned set? We study this question in the following manner. We first determine which growing strategy has the smallest validation loss at step 600,000, and we let $G^\star$ denote this growth operator. Then, at every step $i \in \{500000,\dots,500200\}$, we identify the $k$ growth operators with the smallest validation loss and ask whether $G^\star$ is among these $k$ growth operators; if it is, we plot 1, and if not, we plot 0. Figure~\ref{fig:bert_12_to_16_deep_dive} (bottom left) displays this plot for $k=1,2,3$. As can be seen in the figure, the recall rises to 1 for all values of $k$ around step 500,050, which aligns with the phase transition identified in the correlation heatmap. This provides further support for the conjecture that the phase transition observed above is a transition into a more stable phase with high predictive power with respect to the optimal growing strategies.

\looseness-1\textbf{Notion of regret.}~~~Beyond just identifying or failing to identify the optimal growing strategy at a particular step is the more nuanced question: \textit{how} suboptimal (with respect to validation loss) is it to choose a growing strategy based on its performance at step $i$ (vs. its performance at the end of training). Let $G^i$ denote the growth operator with the smallest validation loss at step $i$, and let $\ell(G)$ denote the final validation loss at step 600,000 after growing with operator $G$ at step 500,000. Among all growing strategies in the search space, let $\ell^\text{min}$ denote the smallest validation loss at step 600,000, and let $\ell^\text{max}$ denote the largest validation loss at step 600,000. Then, for every step $i \in \{500000,\dots,500200\}$, we can calculate the regret as
\begin{equation}\label{eq:regret}
\ell(G^i)-\ell^\text{min}
\end{equation}
and the relative regret as
\begin{equation}\label{eq:relative_regret}
(\ell(G^i)-\ell^\text{min}) / (\ell^\text{max}-\ell^\text{min}).
\end{equation}
In other words, the relative regret captures how suboptimal it is to choose a growing strategy based on its validation loss at step $i$. Figure~\ref{fig:bert_12_to_16_deep_dive} (bottom right) displays the relative regret for $i \in \{500000,\dots,500200\}$. We can see that the relative regret starts at approximately 0.8 and drops to 0 around step 500,050, which aligns with the phase transition identified in the correlation heatmap. This provides further support for the conjecture that the phase transition observed above is a transition into a more stable phase with high predictive power with respect to identifying low-loss strategies.

\looseness-1\textbf{Conclusion.}~~~Taken together, these results demonstrate that, soon after growing, (1) an approximate ordering of the growing strategies emerges and (2) it is possible to identify a low-regret strategy.

\section{Applications}
\label{sec:applications}

\looseness-1In this section, we extend the insights of Section~\ref{sec:bert_exps} into algorithms for growing and stacking. Our goal is primarily to validate the landscape-aware theory through its algorithmic utility when applied in a very simple manner. We believe that more sophisticated algorithms could be developed to further exploit the landscape-aware theory, with even stronger performance, and that this is merely evidence of a step in the right direction algorithmically.

\subsection{LAG}

\looseness-1We define Landscape-Aware Growing, or LAG@$k$, as a simple algorithm for growing as follows. Consider a design space $\mathcal{G}$ of growth operators. For each growth operator $G \in \mathcal{G}$, apply $G$ to the pretrained model $M_1$ to obtain a larger model $M_2$. Train each such $M_2$ for $k$ steps, for some small $k$. Choose the growth operator $\hat{G}$ yielding the lowest validation loss at $k$ steps, and then train $\hat{G}(M_1)$ to completion. Although LAG@$k$ is fairly generic and can be instantiated with various values of $k$, for the purposes of this evaluation, we look for a phase transition as in Figure~\ref{fig:bert_12_to_16_deep_dive} (top left) and choose $k$ to ensure some margin post-phase-transition.

\begin{table}[!tbp]
    \centering
    \caption{Performance of LAG@200 compared to other growing strategies, when growing BERT from 12 layers to 16 layers. See equations~\ref{eq:regret} and \ref{eq:relative_regret} for the definitions of regret and relative regret.}
    \label{tab:bert_growing}
    \begin{tabular}{|c|c|c|c|}
        \hline
         \textbf{Strategy} & \textbf{Final Validation Loss} & \textbf{Regret} & \textbf{Relative Regret} \\ \hline\hline
         Oracle & 1.7461 & 0 & 0 \\ \hline
         LAG@200 & 1.7461 & 0 & 0 \\ \hline
         Best at initialization (LAG@0) & 1.7875 & 0.0414 & 0.7986 \\ \hline
         Stack last block on top & 1.7747 & 0.0286 & 0.5517 \\ \hline
         Stack random block on top & 1.7875 & 0.0414 & 0.7986 \\ \hline
    \end{tabular}
\end{table}

\looseness-1\textbf{BERT.}~~~We first apply LAG to the BERT-\textsc{Base} setting defined in Section~\ref{sec:bert_exps_pitfalls}. Given the phase transition around step 500,050, we use LAG@200 (though smaller values of $k$ would have similar behavior here, based on the results in Figure~\ref{fig:bert_12_to_16_deep_dive}).

\looseness-1In Table~\ref{tab:bert_growing}, we compare the performance of LAG@200 with several other methods. The ``Oracle'' strategy refers to the best possible strategy within the search space, based on validation loss at step 600,000 (i.e., it is not a practical strategy but rather represents the best one could hope to achieve). We also compare to LAG@0; since the loss of $G(M_1)$ is higher than the loss of $M_1$ for all growth operators $G$, choosing the growth operator with the lowest loss after 0 steps of training follows the ``loss preservation'' heuristic (i.e., choosing the strategy whose loss is least perturbed by growing). We also compare to the variants within our search space which most resemble gradual stacking \citep{reddi2023efficient}: (1) stacking the last block on top, and (2) stacking a randomly-initialized block on top. Due to the limited effect of new final layers with small random initialization, a randomly-initialized block on top can be viewed as another proxy for the ``loss preservation'' heuristic and does turn out to match LAG@0.

\looseness-1Overall, with LAG@200, we see that we are able to identify the best-performing strategy, achieving a relative regret of 0. In contrast, the other strategies have a relative regret of at least 0.5 (and even higher).

\begin{table}[!tbp]
    \centering
    \caption{Performance of LAG@2000 compared to other growing strategies, when growing UL2 from 12 layers to 16 layers. See equations~\ref{eq:regret} and \ref{eq:relative_regret} for the definitions of regret and relative regret.}
    \label{tab:ul2_growing}
    \begin{tabular}{|c|c|c|c|}
        \hline
         \textbf{Strategy} & \textbf{Final Validation Loss} & \textbf{Regret} & \textbf{Relative regret} \\ \hline\hline
         Oracle & 2.1254 & 0 & 0 \\ \hline
         LAG@2000 & 2.1268 & 0.0014 & 0.0895 \\ \hline
         Best at initialization (LAG@0) & 2.1398 & 0.0144 & 0.9202 \\ \hline
         Stack last block on top & 2.1357 & 0.0103 & 0.6582 \\ \hline
         Stack random block on top & 2.1398 & 0.0144 & 0.9202 \\ \hline
    \end{tabular}
\end{table}

\begin{figure}[!tbp]
    \centering
    \begin{subfigure}{0.45\textwidth}
    \includegraphics[width=\textwidth]{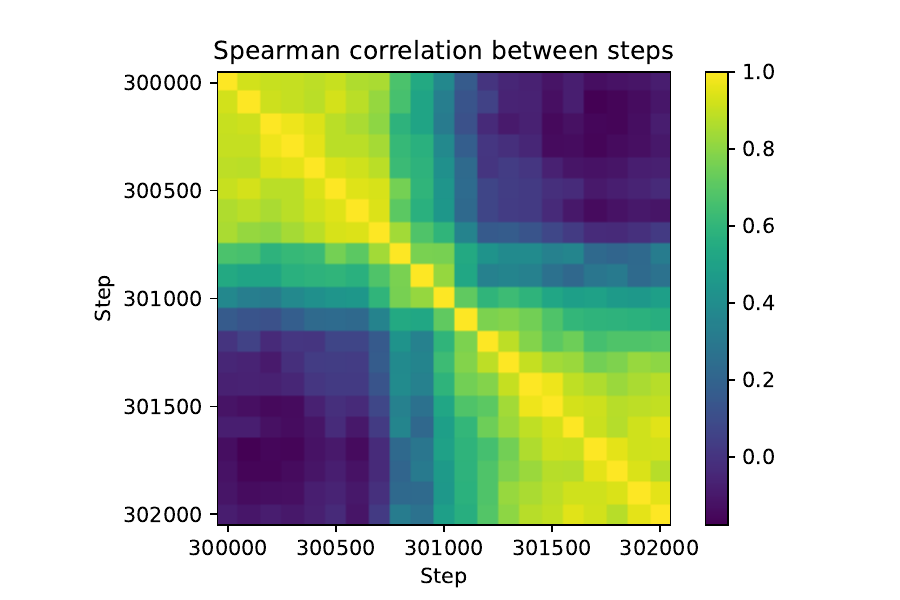}
    \end{subfigure}
    \begin{subfigure}{0.45\textwidth}
    \includegraphics[width=\textwidth]{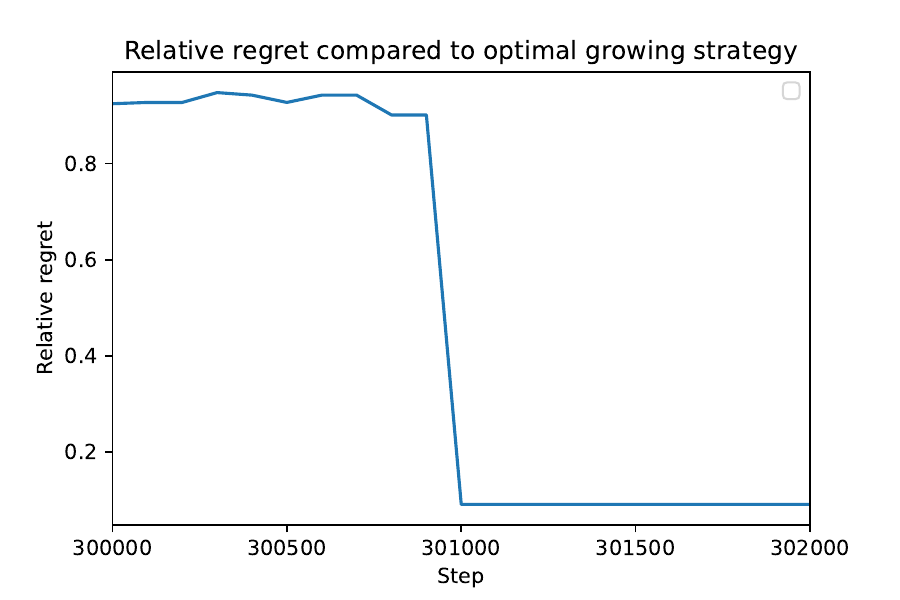}
    \end{subfigure}
    \caption{Growing UL2 from 12 layers to 16 layers. Spearman correlation heatmap (left) and Spearman correlation with final values (right). Here, the validation loss is only measured every 100 steps, so these plots do not use smoothing (in contrast with Figure~\ref{fig:bert_12_to_16_deep_dive}).}
    \label{fig:ul2_heatmap}
\end{figure}

\looseness-1\textbf{UL2.}~~~We also extend LAG to the UL2 setting. Specifically, we begin by pretraining a 12-layer decoder-only model with 1.1B parameters and the UL2 objective \citep{tay2022ul2} for 300,000 steps, before growing it to 16 layers and training for an additional 100,000 steps. See further details in Appendix~\ref{app:training_details}. Since we are growing from 12 to 16 layers as we did in the BERT setting, we include the same 30 growth operators that we studied for BERT in Section~\ref{sec:bert_exps_pitfalls}. We compute the correlation heatmap as defined in Section~\ref{sec:bert_exps_hundredsteps} over the first 2,000 steps immediately after growing (i.e., step 300,000 through step 302,000) and observe a phase transition around step 1,000. To ensure a margin post-phase-transition, we choose $k=2000$, i.e., LAG@2000. In Table~\ref{tab:ul2_growing}, we compare the performance of LAG@2000 with the same alternative methods we examined for BERT, above. Overall, we see that LAG@2000 achieves a relative regret of 0.0895, which is fairly small and much smaller than that of the alternative methods. We also include 1-shot evaluations for several key downstream tasks in Appendix~\ref{app:empirical}, providing evidence that the trends in validation loss hold for downstream metrics as well.

\subsection{Adaptive stacking}
\label{sec:stacking}

Here, we consider how to apply LAG to gradual stacking. Specifically, LAG motivates an \emph{adaptive} strategy for gradual stacking: at each stage of stacking, multiple growing strategies are spawned in parallel and trained for $k$ steps, for some small $k$. Then, the strategy with the lowest validation loss is chosen and training is continued for the rest of the stacking stage using just this one strategy. Applied iteratively, with $n$ stages and $s$ strategies explored per stage, this costs $n \times k \times s$ additional steps of training (divided over a range of model sizes). Here, we use adaptive stacking to train a 24-layer BERT-{\sc large} model in 6 stages, using a roughly uniform stacking schedule (160,000 steps per stage for the first 5 stages, and 200,000 steps in the final stage). We use $k=200$ steps. In Table~\ref{tab:adaptive_stacking} in Appendix~\ref{app:empirical}, we see that adaptive stacking outperforms last stacking \citep{reddi2023efficient}, with a final validation loss of 1.5301 for adaptive stacking vs. 1.5432 for last stacking (i.e., post-stacking in \cite{reddi2023efficient}).

We caveat that this is merely intended as a demonstration of how one might naively apply the principle of LAG to stacking. Without further iteration, we can already see gains over fixed stacking, suggesting that this could be a promising direction for future work on improving stacking (or ``iterated growing'', more generally).

\section{Related work}
\label{sec:related}

The literature on \emph{growing models} is vast; hence, we only focus on the most relevant works here.
Net2Net \citep{chen2016net2net} was one of the first works to popularize parameter-reusing growth operators for neural networks, though notably building on much earlier works such as \citep{cascadecorrelation1989} and \citep{knowledgetransfer2008}. The primary technical contribution of \citep{chen2016net2net} is its \textit{function-preserving} growth operators, which ensure that the new, larger network at initialization represents exactly the same function as the original, smaller network. Specifically, among its main contributions, \citet{chen2016net2net} highlights how function-preserving growth operators avoid ``spending time passing through a period of low performance''. Since 2016, various works have built upon this \emph{function-preserving} idea, and it has become a core tenet in the design of neural network growth operators \citep{weimorphism2016,evci2022gradmax}. 

More recent works in this direction are largely based on Transformers. \cite{bert2bert}, \cite{shen2022staged}, \cite{li2023flm}, \cite{yao2024masked}, and \cite{wang2024lemon} all explore different variants of function- or loss-preserving growth operators specifically in the Transformer setting. Despite differences in how they achieve function preservation, they are all united by the perspective - dating back to Net2Net - that ``a growth operator that is not loss-preserving wastes time and compute initially until it recovers the same performance of the original model'' \citep{wang2024lemon}. It is this design principle that our work provides compelling evidence against.

Although the primary focus of our work is how to expand a shallow model into a deeper model through one step of growing, we also explore its application to iterated growing. Our work is thus related to the literature on progressive stacking \citep{gong2019efficient} and gradual stacking \citep{reddi2023efficient}, which gradually increase the model depth in stages by reusing layers from the previous stage.

\section{Discussion}
\label{sec:discussion}

\looseness-1\textbf{Conclusion.}~~~Overall, we have conducted a fairly extensive empirical analysis of various growth operators and identified that, despite a vast body of prior work on function- and loss-preserving growth operators, initial loss immediately upon growing is not particularly predictive of final performance. Rather, allowing a growth operator to initially disrupt the function (and thus the loss) can actually be desirable if it then leads to a more favorable early loss landscape in which the loss can decrease more rapidly. To that end, we identify that this notion of “early loss landscape” is actually quite early in BERT and is delineated by a measurable phase transition. Based on these insights, we introduce Landscape-Aware Growing (LAG) as a general, design-space-agnostic strategy for growing; with just a little “lag” after initialization, identifying a low-regret growth operator is possible. We validate our approach in UL2 and extend our insights to stacking as well.

\looseness-1\textbf{Limitations.}~~~Although these results are quite exciting, we find it prudent to point out various limitations beyond the caveats already noted. One limitation is that, due to compute constraints, we could only run one trial per growth operator. We hope that by validating our results in a different, more complex setting (UL2), we mitigate these concerns a bit; however, in an ideal world, we would have multiple trials with different random initializations and different data orders. Another limitation is that we have only explored the BERT and UL2 settings, and our largest model sizes are just over 1B parameters; thus, we do not know how our results generalize to state-of-the-art model sizes. Finally, due to compute constraints, our search space is naturally limited to a subset of all possible growth operators. We have tried to capture a range of both approximately loss-preserving and loss-disrupting growth operators, and we thus believe our search space is quite reasonable with respect to our conceptual insights; however, future work could consider expanding the search space to include even more growth operators from the literature.

\looseness-1\textbf{Broader impact.}~~~Although our work is primarily analytical in nature, we do believe it takes an important conceptual step forward in the development of ever-larger language models through the growing of previously-trained models. Therefore, significantly downstream of this work, we see potential positive societal impacts from increased model capabilities at ever-larger scales, as well as potential negative societal impacts if such highly capable models are not developed safely.

\bibliography{references}
\bibliographystyle{plainnat}

\appendix

\newpage
\section{Training details}
\label{app:training_details}

\subsection{BERT growing.} We train BERT-{\sc base} on Books and Wikipedia following \cite{devlin2018bert}. We use batch size 256 and sequence length 512, AdamW as the optimizer \citep{loshchilov2018decoupled}, 10,000 steps of linear learning rate warmup, and a constant learning rate of 0.0001 following warmup. We train BERT-\textsc{Base} for a total of 500,000 steps and then grow it from 12 layers to 16 layers. We ensure that all models see the data in the same order, to ensure that this does not impact relative performance.

\subsection{UL2 growing.} 

Our pretrained UL2 model is a 12-layer decoder-only model with model dimension 2048, hidden dimension 5120, and 32 attention heads. We train this model using the UL2 objective \citep{tay2022ul2} with 60\% causal LM, 20\% prefix LM, and 20\% span corruption. We train on a mixture of C4 (57\%) \citep{2020t5}, Wikipedia (17\%) \citep{wikidump}, GitHub (17\%), and Arxiv (9\%). We use AdaFactor \citep{shazeer2018adafactor} as the optimizer, with batch size 256, sequence length 1280, and 10,000 steps of linear warmup to a peak learning rate of 0.01, followed by a cosine decay schedule so that step 400,000 would end with learning rate 0.001. However, we pause training at 300,000 steps and then grow the 12-layer model to 16 layers, after which we train for another 100,000 steps (totaling 400,000 steps). We ensure that all models see the data in the same order, to ensure that this does not impact relative performance.

\subsection{BERT stacking.} Here, we present the training details for last stacking and adaptive stacking. Broadly, we follow the stagewise paradigm of \cite{reddi2023efficient}. We train BERT-{\sc large} on Books and Wikipedia following \cite{devlin2018bert}. We use batch size 256 and sequence length 512, AdamW as the optimizer \citep{loshchilov2018decoupled}, 10,000 steps of linear learning rate warmup, and a constant learning rate of 0.0001 following warmup. We train for a total of 1,000,000 steps divided among 6 stages. The first 5 stages have 160,000 steps each, and the final stage has 200,000 steps. In each stage, the total number of layers is increased by 4.

\paragraph{Last stacking.} In each stage, we duplicate the \textit{final} 4 layers and stack them at the end of the network (i.e., ``post-stacking'' from \cite{reddi2023efficient}).

\paragraph{Adaptive stacking.} In each stage, we consider a set of growth operators as in Section~\ref{sec:problem_setup}. To keep the design space more manageable, we restrict the design space to parameter duplication (i.e., no random initialization, beyond the very first stage). We train with each growth operator for 200 steps (i.e., LAG@200) before choosing the best-performing growth operator according to its validation loss and then continuing training with just this operator for the rest of the stage. The set of growth operators considered when initializing each stage is as follows:
\begin{itemize}
    \item Stage 1 (4 layers): random initialization (no existing layers to use)
    \item Stage 2 (8 layers): indices $\{0\}$, block sizes $\{1,2,4\}$
    \item Stage 3 (12 layers): indices $\{0,1,2,3,4\}$, block sizes $\{1,2,4\}$
    \item Stage 4 (16 layers): indices $\{0,1,2,3,4,5,6,7,8\}$, block sizes $\{1,2,4\}$
    \item Stage 5 (20 layers): indices $\{0,2,4,6,8,10,12\}$, block sizes $\{1,2,4\}$
    \item Stage 6 (24 layers): indices $\{0,2,4,6,8,10,12,14,16\}$, block sizes $\{1,2,4\}$
\end{itemize}
Note that we switch from every index to only even indices at Stage 5, in order to keep the search space more manageable.

\newpage

\section{Further empirical results}
\label{app:empirical}

Here, we present downstream metrics for UL2 and the adaptive stacking vs. last stacking results in table form.

\begin{table}[!th]
    \centering
    \caption{Performance of LAG@2000 compared to other growing strategies, when growing UL2 from 12 layers to 16 layers, on downstream evaluations.}
    \begin{tabular}{|c|c|c|c|}
        \hline
         \textbf{Strategy} & \textbf{TyDi QA (en)} & \textbf{Trivia QA} & \textbf{LAMBADA}\\
         & exact match - 1 shot & exact match - 1 shot & accuracy - 1 shot\\ \hline\hline
         Oracle & 28.18 & 13.84 & 8.32 \\ \hline
         LAG@2000 & 25.23 & 13.44 & 7.60 \\ \hline
         Best at initialization (LAG@0) & 21.82 & 12.19 & 6.89 \\ \hline
         Stack last block on top & 21.59 & 12.39 & 6.90 \\ \hline
         Stack random block on top & 21.82 & 12.19 & 6.89 \\ \hline
    \end{tabular}
    \label{tab:ul2_growing_downstream}
\end{table}

\begin{table}[!th]
    \centering
    \caption{Final validation loss of adaptive stacking vs. last stacking.}
        \begin{tabular}{ | c | c | c | c | c |}
         \hline
         \textbf{Strategy} & \textbf{Final validation loss}\\ \hline
         Last stacking & 1.5432\\ \hline
         Adaptive stacking & 1.5301\\ \hline
        \end{tabular}
    \label{tab:adaptive_stacking}
\end{table}

\end{document}